\documentclass[letterpaper, 10 pt, conference]{ieeeconf} 
\IEEEoverridecommandlockouts      
\usepackage{times}
\usepackage[pdftex]{graphicx}
\usepackage{graphics}
\usepackage[cmex10]{amsmath}
\usepackage{url}
\usepackage{array}
\usepackage{tabularx}
\usepackage{multirow}
\usepackage{multicol}
\usepackage{booktabs}
\usepackage{epstopdf}
\usepackage{rotating}
\usepackage{csquotes}

\usepackage{amssymb}
\usepackage{amsmath}
\usepackage[all]{xy}    
\usepackage{ifthen}
\usepackage[usenames,dvipsnames]{color}
\usepackage{enumerate}
\usepackage{bm}
\usepackage{siunitx}
\usepackage{bbm}
\usepackage{xargs}                      
\usepackage[pdftex,dvipsnames]{xcolor}  
\usepackage[colorinlistoftodos,prependcaption,textsize=tiny]{todonotes}
\newcommandx{\unsure}[2][1=]{\todo[linecolor=red,backgroundcolor=red!25,bordercolor=red,#1]{#2}}
\newcommandx{\change}[2][1=]{\todo[linecolor=blue,backgroundcolor=white!25,bordercolor=blue,fancyline,#1]{#2}}
\newcommandx{\info}[2][1=]{\todo[linecolor=OliveGreen,backgroundcolor=OliveGreen!25,bordercolor=OliveGreen,#1]{#2}}
\newcommandx{\improvement}[2][1=]{\todo[linecolor=Plum,backgroundcolor=Plum!25,bordercolor=Plum,#1]{#2}}
\newcommandx{\thiswillnotshow}[2][1=]{\todo[disable,#1]{#2}}

\makeatletter
\let\NAT@parse\undefined
\makeatother
\usepackage[numbers]{natbib}

\newcommand{\secref}[1]{Section~\ref{#1}}
\newcommand{\tabref}[1]{Table~\ref{#1}}

\newcommand{\figref}[1]{Figure~\ref{#1}}

\newcommand{\probref}[1]{Problem~\ref{#1}}

\newcommand{\argmin}{\operatornamewithlimits{argmin}}

\newcommand{\myparagraph}[1]{\vspace{0.1in}\noindent\textbf{#1}}

\newboolean{draft-mode}
\setboolean{draft-mode}{true}
\newcommand{\sidenote}[1]{\ifthenelse{\boolean{draft-mode}}{\marginpar{\tiny\raggedright\textsf{\hspace{0pt}#1}}}{}}
\DeclareRobustCommand{\pynote}[1]{\ifthenelse{\boolean{draft-mode}}{\textcolor{green}{\textbf{PY: #1}}}{}}
\DeclareRobustCommand{\arnote}[1]{\ifthenelse{\boolean{draft-mode}}{\textcolor{blue}{\textbf{AR: #1}}}{}}
\DeclareRobustCommand{\nfnote}[1]{\ifthenelse{\boolean{draft-mode}}{\textcolor{red}{\textbf{NF: #1}}}{}}
\DeclareRobustCommand{\mbnote}[1]{\ifthenelse{\boolean{draft-mode}}{\textcolor{cyan}{\textbf{MB: #1}}}{}}

\newcommand{\bp}{{\bf p}}
\newcommand{\bq}{{\bf q}}

\newcommand{\bx}{{\bf x}}
\newcommand{\be}{{\bf e}}

\newcommand{\br}{{\bf r}}

\newcommand{\bw}{{\bf w}}
\newcommand{\bz}{{\bf z}}

\title{\LARGE \bf Realtime State Estimation with Tactile and Visual Sensing for Inserting a Suction-held Object
}
\author{\authorblockN{Kuan-Ting Yu$^1$, Alberto Rodriguez$^2$} \authorblockA{$^1$
    Computer Science and Artificial Intelligence Laboratory ---
    Massachusetts Institute of Technology\\
    $^2$ Mechanical
    Engineering Department --- Massachusetts Institute of Technology\\
    {\tt\small peterkty@csail.mit.edu}, {\tt\small
      albertor@mit.edu}} \thanks{This work was supported by
    NSF award [IIS-1427050] through the
    National Robotics Initiative and the Toyota Research Institute.}
}

\begin{document}

\maketitle


\begin{abstract}

We develop a real-time state estimation system to recover the pose and contact formation of an object 
relative to its environment. 
In this paper, we focus on the application of inserting an object picked by a suction cup into a tight space, an enabling technology for robotic packaging. 

We propose a framework that fuses force and visual sensing for improved accuracy and robustness. 
Visual sensing is versatile and non-intrusive, but suffers from occlusions and limited accuracy, especially for tasks involving contact. Tactile sensing is local, but provides accuracy and robustness to occlusions. 
The proposed algorithm to fuse them is based on iSAM, an on-line optimization technique, which we use to incorporate kinematic measurements from the robot, contact geometry of the object and the container, and visual tracking.
In this paper, we generalize previous results in planar settings to a 3D task with more complex contact interactions.
%
%
A key challenge in using force sensing is that we do not observe contact point locations directly. We propose a data-driven method to infer the contact formation, which is then used in real-time by the state estimator. We demonstrate and evaluate the algorithm in a setup instrumented to provide groundtruth.

\end{abstract}

\section{Introduction}

\label{sec:introduction}

\begin{figure}
  \begin{center}
  \includegraphics[width=0.8\linewidth]{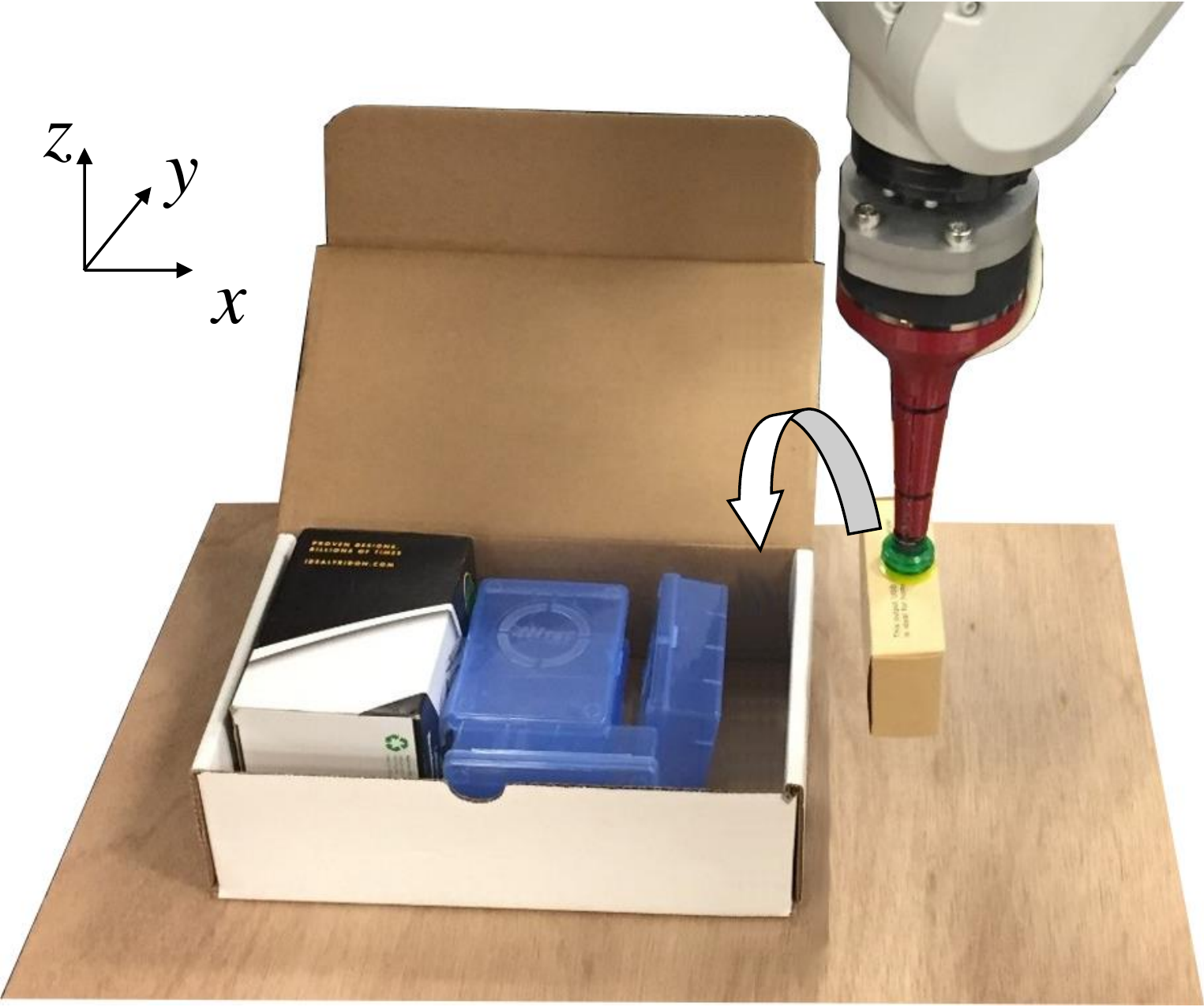}\\
  
    \includegraphics[width=1\linewidth]{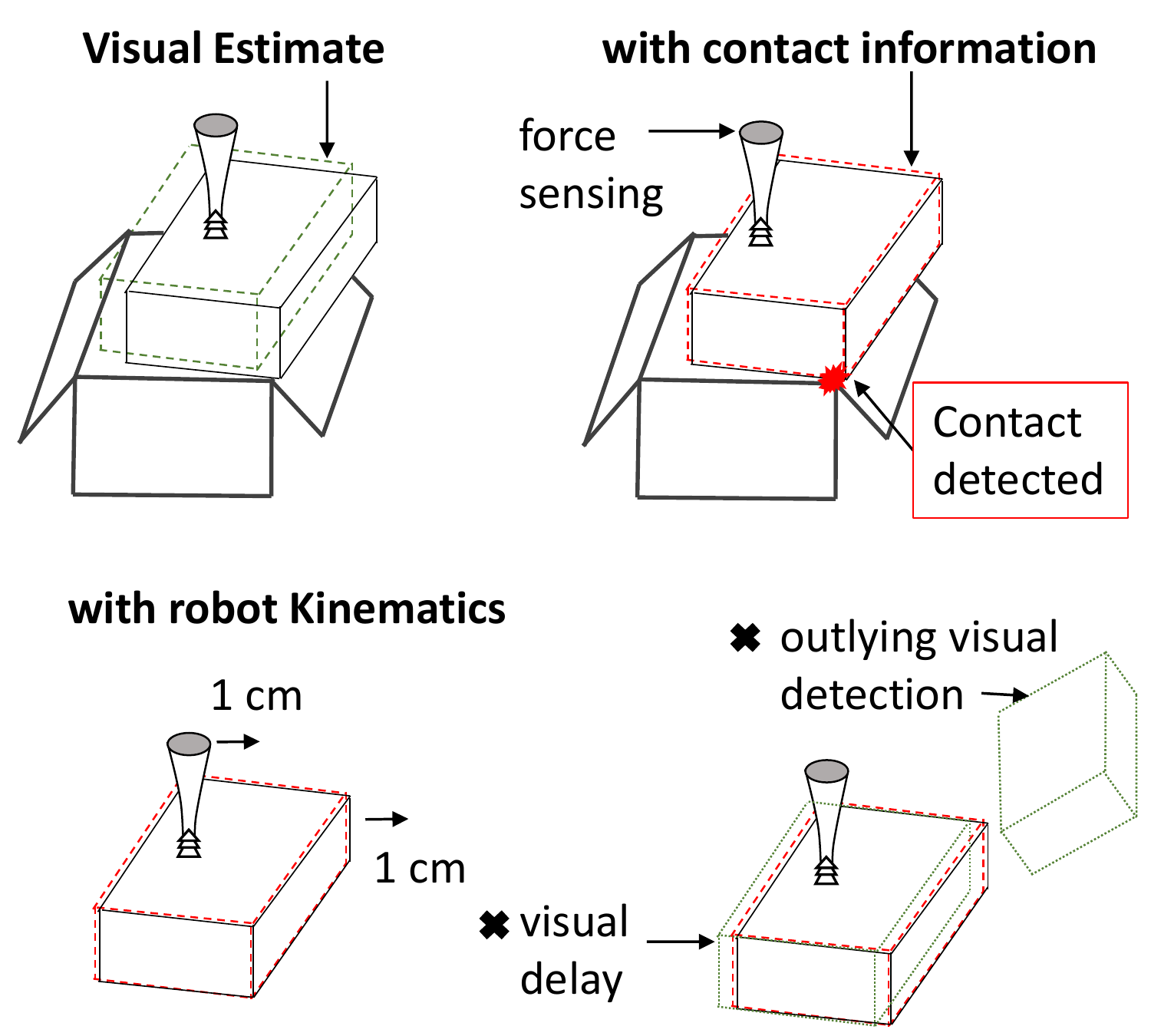}
  \end{center}
  \caption{Concept illustration. (top) Example scene of inserting a suction-held box. (middle) The idea of state estimation with tactile and visual sensing. Green dots: visual estimate; red dashed: visual + tactile estimate; black: groundtruth. After the robot detects the contact formation (i.e., how the object makes contact with the container), the estimate is updated based on the known geometric information. (bottom) The object should move according to the robot motion. This information helps to remove outlying visual detections and delayed visual measurements because of occlusions or latency in visual tracking.}
  \label{fig:scenario}
\end{figure}

We develop a state estimator for the application of inserting an object held by a suction cup into a tight container. This scenario takes inspiration from manual packaging stations in e-commerce warehouses. 
Warehouse bin-picking has been widely addressed in recent work~ \citep{zeng2017robotic} and is close to commercial use~\citep{wsj}. Fine placement of the picked object is a task that has received significantly less attention. In the logistics industry, placing objects densely reduces the space required to store and transport. In this paper, we focus on the state estimation problem for the task of \emph{inserting} an object into a container that is already populated.

The proposed state estimation framework is based on incremental Smoothing and Mapping (iSAM) \cite{Kaess08tro}, an online least-squares optimization method that can add variables and constraints in an on-line fashion and maintain an estimate in real-time. 

We derive a metric that combines observation models for both visual and tactile sensing. Visual sensing provides global information but is noisy and suffers from occlusions and unexpected outlying detections. On the other hand, force/tactile sensing is local but accurate when contact occurs; i.e., the distance is zero. 
Our focus is on exploiting tactile information. In our scenario, the tactile information is provided through a combination of a force-torque sensor measurement at the wrist of the robot, the geometry of the container and the object, and the measurement of robot encoders. The general idea is illustrated in \figref{fig:scenario}.

The state estimation framework extends from our previous work~\citep{yu2018state} on tracking the pose of an object pushed in the horizontal plane. In this paper, we address four new technical challenges:
\begin{itemize}
    \item There are more complex contact formations.
    \item The contact formations cannot be discriminated directly.
    \item The robot has a deformable joint whose state is not directly observable.
    \item The object state is in 3D.
\end{itemize}
The first two contributions, referring to contact formations (CFs), are the foci of this paper. CFs describe the structure of possible contact arrangements, i.e., which edge contacts which edge. To detect CFs, we train a support vector machine (SVM)~\cite{chang2011libsvm, cortes1995support} to distinguish the contact formations given force measurements. We apply a self-supervised scheme to collect labeled examples of contact formations: we use a precise Vicon tracking system that provides enough positional accuracy to estimate contact formation and make use of force sensing to distinguish between contact and no contact.
This self-supervised scheme allows a large-scale data collection that ultimately yields sufficient classification performance. After the contact formations are predicted we add the corresponding cost functions to enforce the contact information.

We show the efficacy of fusing contact and visual information with ablation studies on the performance of the system using different combinations of sensor information, and for two different objects. The instrumented setup is used for the purpose of evaluating the recovered object pose and contact locations.

\section{Related Work}
\label{sec:relatedWork}

This work aims to combine the classical work in peg-in-hole assembly and recent breakthroughs in state estimation techniques for solving the SLAM problem. 
\citet{gadeyne2005bayesian} show that there is an analogy between the SLAM problem and state estimation for the manipulation task involving contacts: both infer continuous variables (poses of the robot v.s. the object) and discrete variables (data association v.s. contact formation detection). Below we discuss the literature of A) the peg-in-hole problem; and B) state estimation in manipulation tasks.

\subsection{Peg-in-hole problem}

The peg-in-hole problem studies the task of inserting an object into a hole where the tolerance is smaller than the accuracy of the actuators or sensors. There are two important families of classical approaches: 
the design of passive compliant devices or control schemes~\citep{drake1978using,whitney1982quasi} which are often tailored to a specific scene; or active sensing techniques with feedback control \citep{simunovia1979information,inoue1974force,bruyninckx1995peg}. The latter provides more potential flexibility in adapting to new scenarios.
In this work, we investigate the second approach, where the quality of sensing and estimation often dictates the system performance. The focus of this paper is on providing high quality and timely estimates of the object pose and contact formation

\citet{simunovia1979information} first presents the insertion process as a pose estimation problem between the part and the hole from a series of noisy measurements. He used both smoothing and Kalman filtering methods. However, due to lack of computation power, the algorithm did not work online. 
We follow his idea and use iSAM to balance accuracy and latency. In addition, we provide an in-depth experimental understanding of the performance of the system.

\citet{bruyninckx1995peg} proposed a model-based state estimation approach for the initial alignment of peg and hole. Their view is close to ours, where the initial alignment is more crucial than the later insertion stage. However, their method is only derived for a round peg and round hole. It is nontrivial to apply it to other geometries. In fact, many of the classical work is only derived for round-peg-round-hole settings in which symmetries simplify the problem. Round profiles are common in manufacturing, but not in a warehouse setting.

The work that exploits passive compliance in hardware or control is well-developed and maybe more suitable for industrial application.
\citet{drake1978using} proposes a passive compliance device called Remote Center Compliance (RCC) for accommodating minor uncertainties in geometry in the insertion task.
\citet{whitney1982quasi} presents analysis to choose the compliance parameters for a given task, and derives conditions to avoid \emph{wedging} and \emph{jamming}.
However, the analyses are planar by assuming a round peg and round hole with chamfers.
\citet{caine1989assembly} extend Whitney's analysis to inserting a rectangular part to a chamferless hole, which is closest to our box inserting scenario, where the suction cup can be seen as a passive compliant device. His analysis of contact formations inspires our work. 

\subsection{State estimation in manipulation tasks}
Probabilistic filters such as particle filters and extended Kalman filters are widely used in robotics~\citep{thrun2005probabilistic}. Particle filters are good for representing complex multimodal distribution that often arises in contact problems, but struggle to provide accurate estimation due to two problems. First, they are subject to sampling resolution. Second, they usually face particle depletion problem when fusing measurements of different levels of noise. \citet{koval2015} proposed a manifold particle filter for handling the depletion problem. The manifold generation is usually done offline for a particular configuration of an end-effector in order to make the estimation fast enough online. 

\citet{gadeyne2005bayesian} propose to solve the joint probability of the hybrid problem (i.e., with continuous and discrete variables) using particle filters. 
In the same line of work, 
\citet{zhang2012application} use nonlinear complimentarity programs to resolve contact and motion simultaneously. However, the expensive computation is far from real-time.
\citet{li2015state} use a contact graph that governs the transition of discrete contact states so that the contact state evolves physically. The system, however, does not scale well due to the combinatorial growth of the contact graph. Our system uses a simple SVM classifier to estimate the instantaneous contact formation parallel to the probabilistic inference. 

The Extended Kalman Filter (EKF) is a popular framework for online state estimation also used in contact tasks~\cite{thrun2005probabilistic,izatt2016gelsight, hebert2011fusion}. It linearizes a system so as to apply a Kalman Filter, designed for linear systems. One common drawback is that the linearization point is at the current estimate of variables. This can result in an inaccurate linearization, followed by an inaccurate estimation.

\citet{Kaess08tro} propose incremental smoothing and mapping (iSAM) to solve the above issues in the SLAM setting. iSAM can be seen as an online nonlinear least-square optimization tool, where cost functions and variables for the optimization can be added online and it can refine the current estimate of the variables and linearization points. The update is efficient because it uses a QR-factorized matrix to represent the linearized cost functions and only updates a small portion of the matrix. 

We use iSAM in our previous work for state estimation in the planar situation \cite{yu2018state} and obtain a positive result in terms of higher accuracy over EKF while considering the realtime constraint. In this work, we explore a similar framework but on a problem with more complex contact formations.

\section{Suction-based insertion problem}

We are concerned with estimating the pose of a suction-held object, its contact formation with a hole, and the precise location of the contact points between them. We assume the object and the hole are rigid. We particularly focus on the mating phase -- aligning of the object and hole -- and lesser on the following pushing-down stage. The mating phase is key to the insertion problem and the contact conformation is richer. 

In a complete assembly system, the output of our algorithm would be fed into an insertion controller. The top of \figref{fig:scenario} shows an example scenario where a box is to be inserted into the remaining free space in a box. 

Based on this application, we simplify the scenario to the experimental setup in \figref{fig:setup} composed of two rigid parallel walls that emulate the space where to insert the box. These walls play the role of either the container or adjacent objects. We also assume only one dimension of the hole is tight not the other dimension (i.e., $x$ versus $y$ direction in \figref{fig:setup}). This is often true for real packing scenarios, and also simplifies the instrumentation and observation of the assemblies.



\subsection{Problem definition}
The object state is observed with
periodicity and we use the subscript $t\in[1...T]$ to indicate the
corresponding timestamp along the trajectory.  

\myparagraph{State variables.} The state $\bx_t$ at time $t$ includes object pose $\bp_t$ and contact points $\bq_t$. The object pose is denoted as $\bp_t = (x,y,z,\phi,\theta,\psi)$, where the latter three variables are roll, pitch, and yaw, respectively. The pose is with respect to the top center of the hole. 
The contact points on the walls and on the objects are denoted in the object frame as $\bq_t=\{q_i^w, q_i^o\}$, respectively, where $i$ is a unique id. These points will be defined based on the possible contact formations of the objects involved. 

When in contact, the contact points on the objects and the walls should be coincident. Here we parameterize both and will impose their coincidence as a constraint when appropriate depending on the predicted contact formation. This makes the implementation more modular and easy to adapt to changes in the object or hole. \secref{sec:contact} describes how to impose contact information in our design.

\myparagraph{Visual input.} Visual input is in the form of time-stamped 3D poses $\bw_t=(x,y,z,\phi,\theta,\psi)$ and a binary signal denoting whether it is available at time $t$ or not (possibly due to occlusions). In this implementation, we use Apriltag markers to emulate a perception system with a realistic noise level. The algorithm, however, is agnostic to the particular type of perception algorithm.

\myparagraph{Tactile input.} A tactile input $\bz_t=\{\br_t, {\bf f}_t\}$ includes the 3D pose of robot's tool center point (TCP) $\br_t=(x,y,z,\phi,\theta,\psi)$ and force-torque input at TCP ${\bf f}_t=(f_x,f_y,f_z,\tau_x,\tau_y,\tau_z)$.

\section{Method}
In this section, we formulate the state estimation problem as a least squares problem in order to apply iSAM. The iSAM algorithm allows adding variables and constraints in an on-line fashion, in contrast to batch optimization. We refer the reader to \citep{Kaess08tro} for a details description of the iSAM algorithm.

\subsection{Objective function}
The overall cost function is a sum of five cost functions/constraints: 
\begin{itemize}
  \item the visual measurement cost $V$; 
  \item the local robot motion cost $M$;
  \item the contact measurement cost $C$; 
  \item the contact point on geometric feature cost $L$;
  \item the contact point prior cost $Q$. 
\end{itemize}

\begin{figure}
  \begin{center}
    \includegraphics[width=0.9\linewidth]{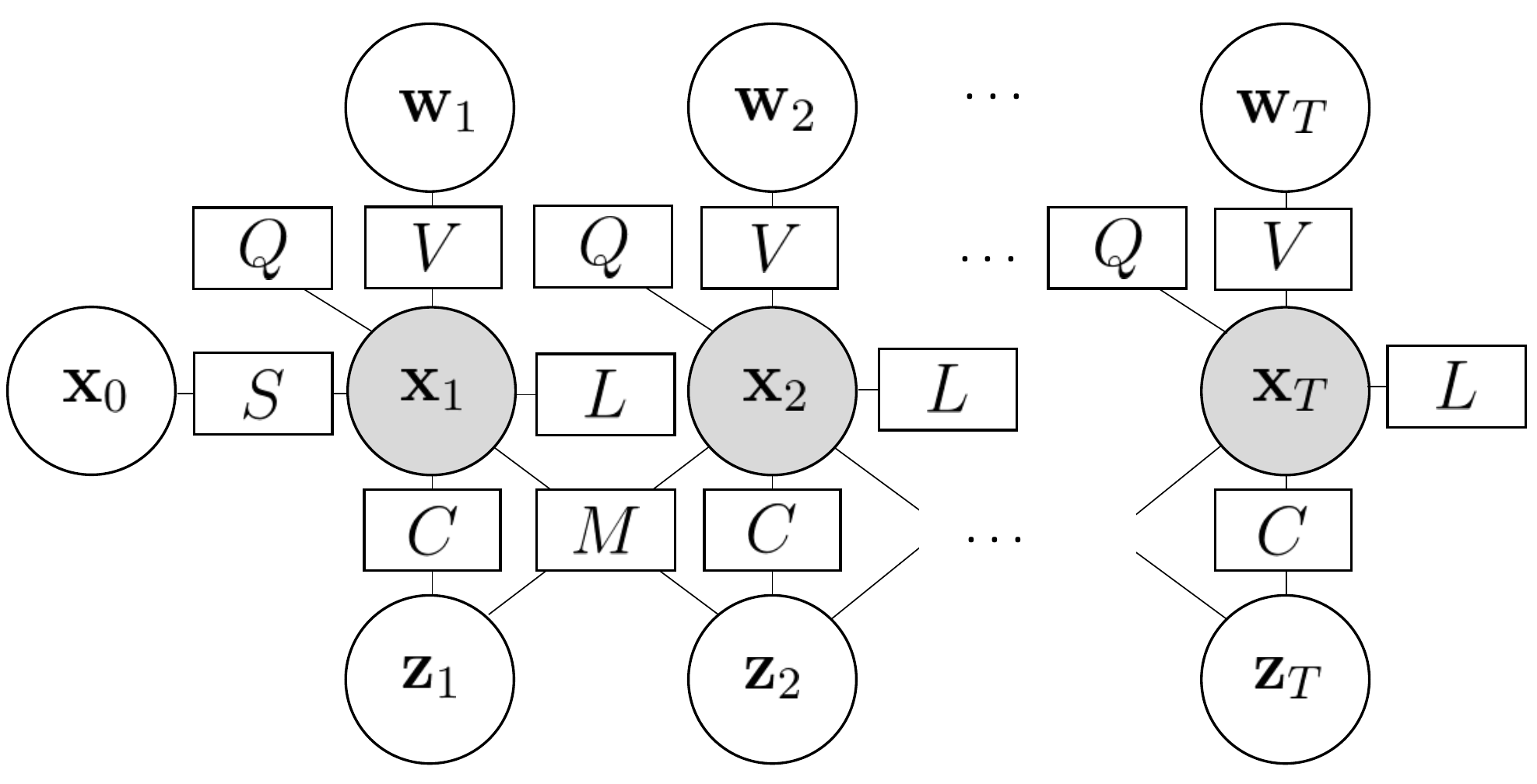}
  \end{center}
  \caption{A factor graph representation of the relationship between the variables and cost functions. The shaded circles represent the state variables to estimate, which are the object poses. The unshaded circles represent sensor measurements including camera inputs (\bw) and tactile inputs (\bz). The rectangles are the cost functions ($V, M, C, L, Q$) that enforce soft constraints between the variables and measurements. }
  \label{fig:factor_graph}
\end{figure}

The factor graph in \figref{fig:factor_graph} shows the relationship between these cost functions. The overall least squares problem is then:
\begin{equation}
\begin{aligned}
\label{eq:obj}
X^* = \argmin_X \sum_{t=1}^T 
&\left \| V(\bx_{t}, \bw_t) \right \|_{\Omega_V}^2  \\
+&\left \| M(\bx_{t-1}, \bx_{t}, \bz_{t-1}, \bz_{t}) \right \|_{\Omega_M}^2 \\
+&\left \| C(\bx_t, \bz_t) \right \|_{\Omega_C}^2 \\
+&\left \| L(\bx_{t}) \right \|_{\Omega_L}^2 \\
+&\left \| Q(\bx_{t}) \right \|_{\Omega_Q}^2 ,
\end{aligned}
\end{equation}
where $X$ is a long vector formed by concatenating ${\bx_t}$'s, and $\left\|\be\right\|_\Omega=\be^T\Omega^{-1}\be$ computes the squared Mahalanobis distance with covariance matrix $\Omega$. The $\Omega$ matrices are the covariance matrices for the corresponding noises of every constraint. We identify them from the measurement input and groundtruth. If some measurement is missing due to physical limitations, we remove the relevant cost functions at that instant; e.g., when the object is not in camera view, we remove the $V$ term.

Below we discuss the visual measurement and motion cost. The contact related costs are discussed in \secref{sec:contact}.

\subsection{Visual measurement $V$}

The visual measurement cost forces the pose estimate to be close to the visual input. The associated constraint is simply the subtraction of both inputs:
\begin{equation}
\begin{aligned}
V({\bf x}_t,{\bf w}_t) & =  {\bf x}_t - {\bf w}_t
\end{aligned}
\end{equation}




\subsection{Motion prior $M$}

The motion prior forces the displacement of the object to be consistent with the robot motion. The deformation of the suction cup is small, and its effect on the object pose is mostly in rotation, so we adjust the noise level to match the stiffness of a suction cup.

This cost can stabilize the estimation to be robust against noise in the visual input. This cost also allows the correction from contact to persist after the contact is broken.
This is implemented with a subtraction of the two displacements: 
\begin{equation}
\begin{aligned}
M({\bf x}_{t-1},{\bf x}_t,{\bf z}_{t-1},{\bf z}_t) & =  ({\bf p}_{t}-{\bf p}_{t-1}) - ({\bf r}_{t}-{\bf r}_{t-1})
\end{aligned}
\end{equation}

All the subtraction operations in cost functions on angles are wrapped into $[-\pi,\pi)$.

\section{Contact information}

\label{sec:contact}

The optimization framework iSAM handles continuous variables. In our implementation, we decide the discrete contact formation (CF) first and then provide iSAM the corresponding constraints. This enables the estimation to be in real time. Below we illustrate the method with the example of a cuboid.

\subsection{Contact formations}

A general guideline for defining CFs is to minimize the number of CFs, since classification accuracy will suffer when there are too many classes.

\begin{figure}
  \begin{center}
    \includegraphics[width=0.7\linewidth]{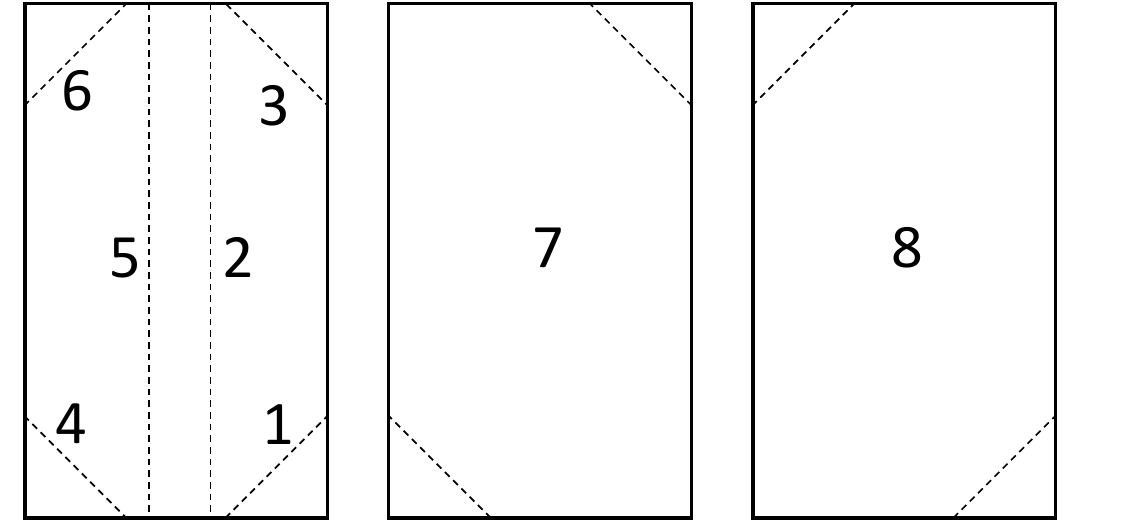}
  \end{center}
  \caption{Contact formations of the rectangle object. The rectangle represents the bottom face of the object. The dashed lines represent the line contacts of modes of 1 to 8 with the top face of the two walls. Mode 0 represents no contact.}
  \label{fig:contact_formations}
\end{figure}

We define 9 contact formations labeled from 0 to 8 as shown in \figref{fig:contact_formations}. The contact formations are line contacts between the bottom face of the object and the top face of the walls.
There are possible degenerate situations which we do not include, like a face contact, which we include in the case of line contact.

We propose to use a support vector machine (SVM) to classify the received force signature during the mating phase of the assembly to estimate CFs. 
The input to the classifier is 6-dimensional force-torque measurement, and the output is a discrete CF id. Based on the hyper-parameter tuning tool provided by libsvm~\citep{chang2011libsvm}, we choose the radial bases function as the kernel of SVM. We choose SVM because of its speed, simplicity, and because it gives reasonable accuracy in our scenario. One could use many other classifiers.

\subsection{Contact constraint $C$}

\begin{figure}
  \begin{center}
    \includegraphics[width=1\linewidth]{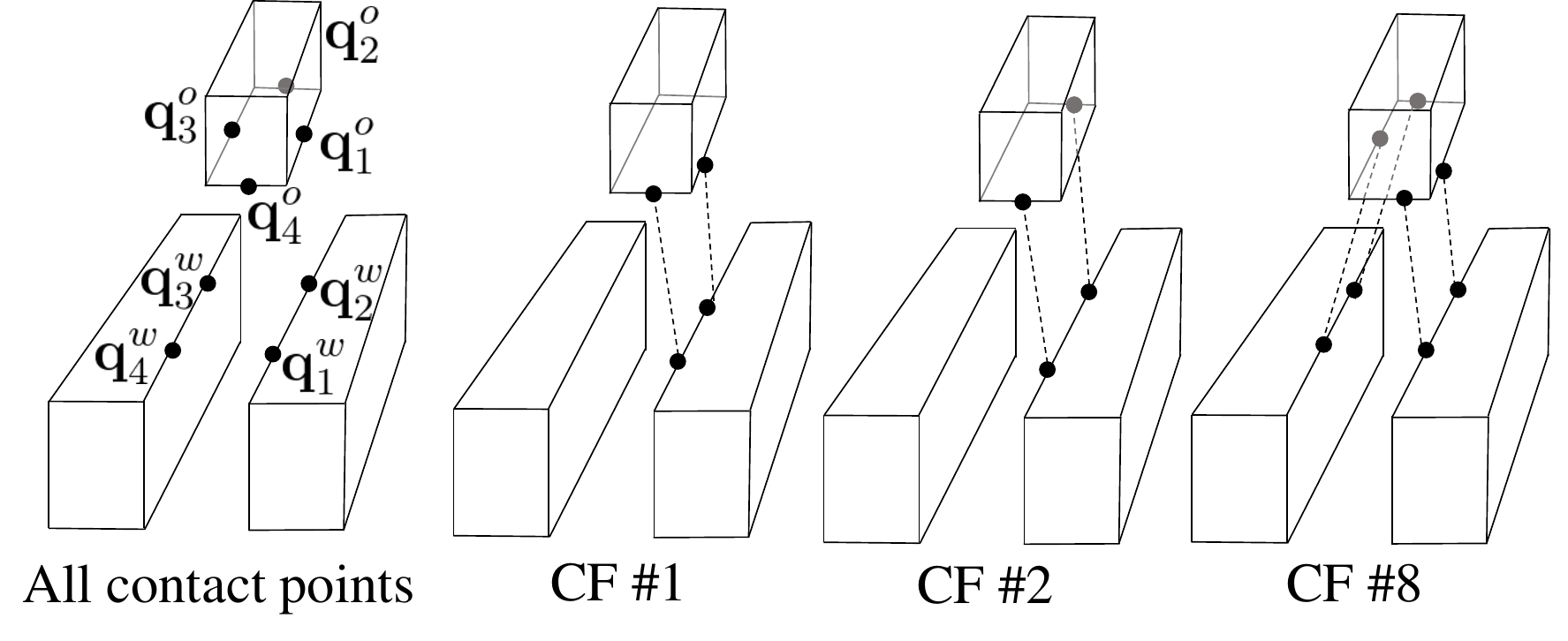}
  \end{center}
  \caption{All contact points and the contact constraints of contact formation 1, 2, and 8.}
  \label{fig:contact_constraint}
\end{figure}

We aim for a modular and simple formulation to describe contact constraints because contact constraints take much effort to implement correctly by enumerating the contact formations. We want it to be easy to adapt to other geometries and to scale as the geometric complexity grows. 

The process for specifying contact constraint is as follows. Regardless of whether contact happens or not, we always define contact points on the objects and walls. We assign 4 contact points on the bottom edges of the object and 2 contact points to the individual walls as shown in \figref{fig:contact_constraint}. When a particular contact formation is detected by the SVM classifier, we \emph{connect} the corresponding contact points to force them to join. \figref{fig:contact_constraint} shows examples of the contact constraints of the \texttt{rect} object.

The contact constraint is implemented as a subtraction of the two points in the wall coordinate. Note that the contact points on the object are in object frame, and contact points on the wall are in the wall frame. Thus, we define

\begin{equation}
\begin{aligned}
C({\bf x}_t) = \{ {^wT_o}{\bf q}_i^o - {\bf q}_j^w \},
\end{aligned}
\end{equation}
for all $\bq_i^o$ and $\bq_j^w$ that meet according to the CF. Here ${^wT_o}$ is a homogeneous transform matrix that transform ${\bf q}_i^o$ in object coordinate to wall coordinate. Note that there could be multiple contacts, so there will be multiple $C$ terms in Eq. \eqref{eq:obj} according to the CF.

\subsection{Contact point on geometric feature $L$}
We explicitly constrain the contact point to be along a designated geometric feature (e.g., line, curve) on a object or a wall. The cost is the difference between the point and its corresponding closest point $\hat{\be}$ on the line/curve. That is,

\begin{equation}
\begin{aligned}
L({\bf x}_t) = \{ \bq_i^w - \hat{\be} \}.
\end{aligned}
\end{equation}

\subsection{Contact point prior $Q$}
The contact point prior is a regularization term that gives a gentle hint of the contact point location on a geometric feature on the object or walls. This prior has two benefits. First, it helps to locate the contact points based on prior statistics. Second, it prevents the system from becoming under-constrained, which will cause the least squares problem ill-defined.
In practice, we use relatively low weight compared to that of other costs.
Therefore, this prior constraint is only effective when there are no other relevant cost. The cost is the difference of the contact point, ${\bf q}_i$ and its corresponding nominal position, $\hat{{\bf q}}_i$:

\begin{equation}
\begin{aligned}
Q({\bf x}_t) =  \{ {\bf q}_i - \hat{{\bf q}}_i \}.
\end{aligned}
\end{equation}
The nominal position can be found by statistics of the groundtruth data. This does not have to be accurate as it is a \emph{regularization} prior.

\subsection{Adaptation to another shape -- elliptic cylinder}
We choose elliptic cylinder to test a geometry with curved shape. Following the above procedure, we can adapt the cost functions to an elliptic cylinder. We mainly need to change three places:
\begin{itemize}
    \item[1.] The definition of the contact formations;
    \item[2.] The associated contact constraints;
    \item[3.] The contact point on the geometric feature of the object, i.e., the elliptic bottom face.
\end{itemize} 

\begin{figure}
  \begin{center}
    \includegraphics[width=0.8\linewidth]{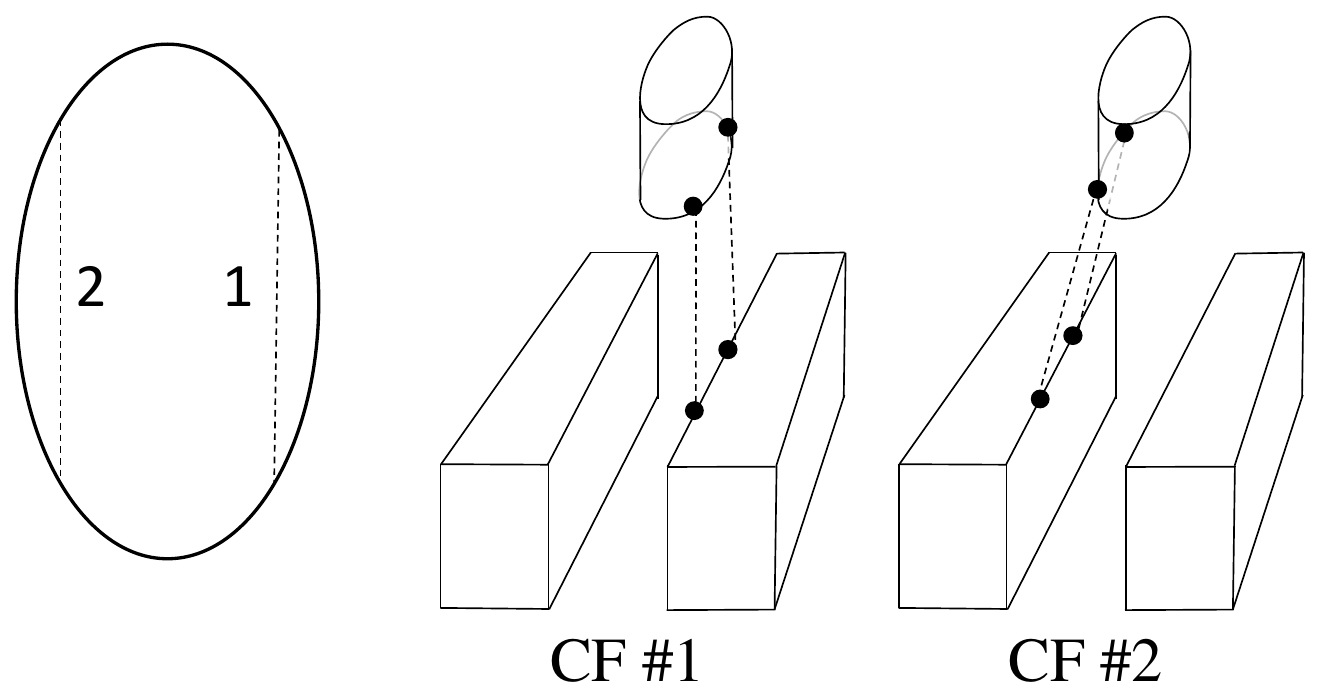}
  \end{center}
  \caption{Contact formations and contact constraints of object \texttt{ellip}. We focus on the bottom part of the real object, which is an elliptic cylinder. We do not expect the upper part of the object to make contact with the wall.}
  \label{fig:contact_formation_ellip}
\end{figure}

The contact formations and contact constraints for \texttt{ellip} are shown in \figref{fig:contact_formation_ellip}. Since the ellipse is a smooth shape, we only need two CFs to describe which wall is in contact. Due to our modular design, we do not need to modify the constraint for contact point on walls. 


\section{Experiments}
\label{sec:experiment}


%
We want to understand the individual components and how possible changes may affect the performance of the system. Specifically, we want to answer the following questions.

\begin{itemize}
  \item How accurate is the contact formation prediction?
  \item How much do contact constraints improve accuracy?
  \item How accurate is the contact point prediction?
  \item How well does the system adapt to an object of smooth shape, e.g., elliptic cylinder?
  \item What is the runtime of the system?
\end{itemize}

\subsection{Setup}
\begin{figure}
  \begin{center}
    \includegraphics[width=1\linewidth]{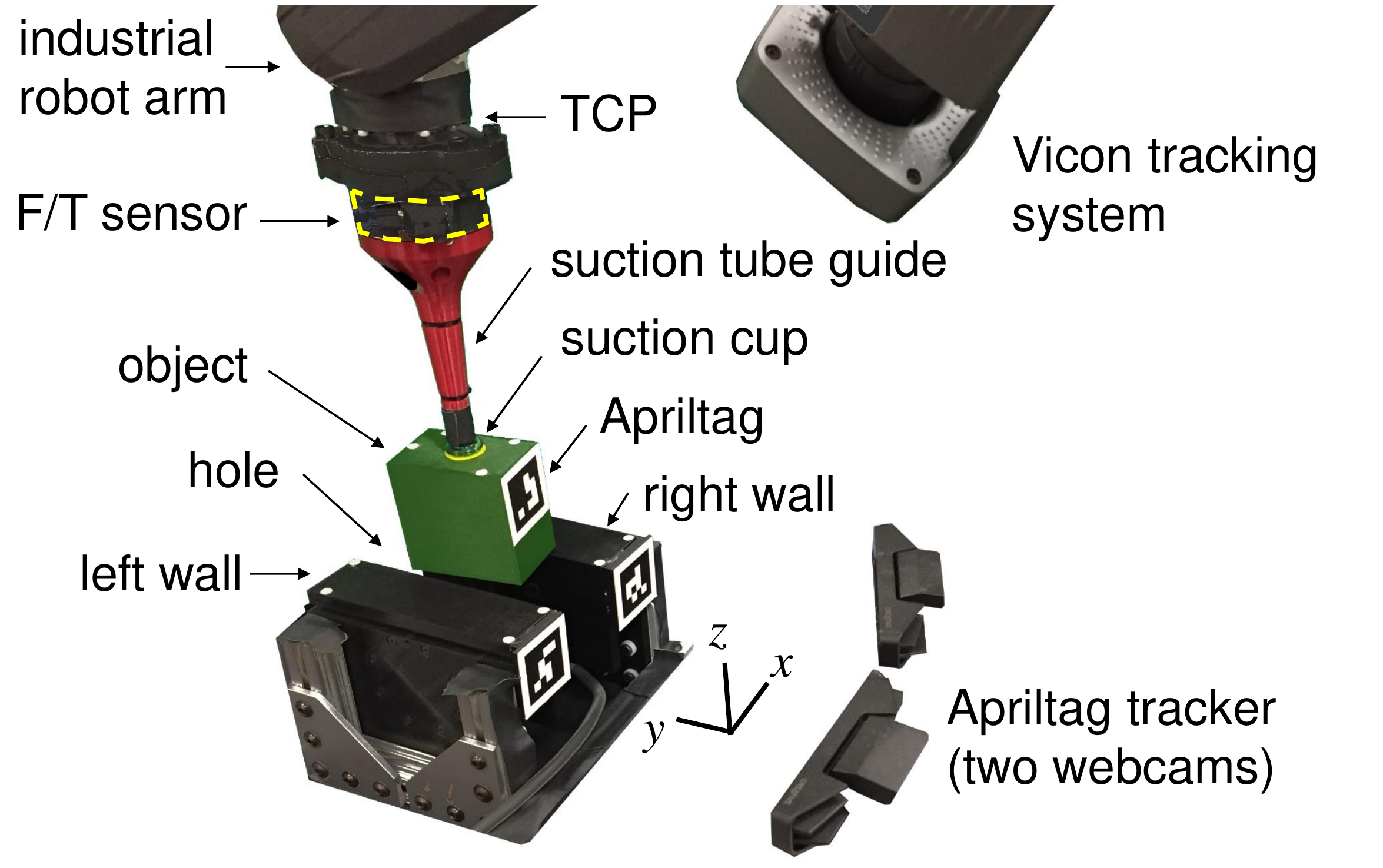}
  \end{center}
  \caption{Experimental setup. }
  \label{fig:setup}
\end{figure}

Our experimental setup is shown in \figref{fig:setup}.
Details about the experimental software, data and video results are available at \url{mcube.mit.edu/ins-est}.

\begin{figure}
  \begin{center}
    \includegraphics[width=0.7\linewidth]{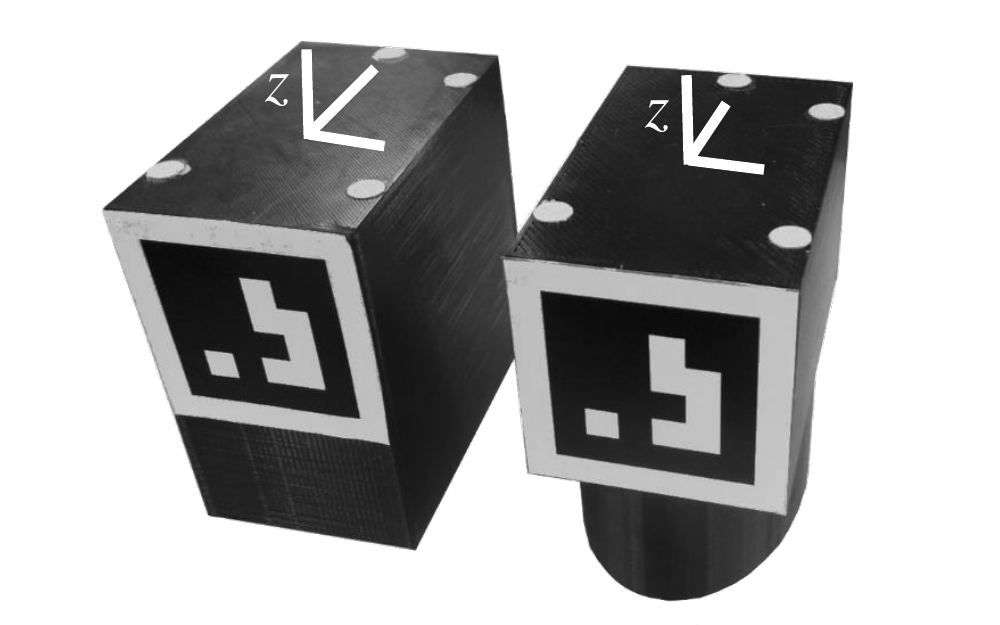}
  \end{center}
  \caption{Experimental objects. (left) \texttt{rect}. (right) \texttt{ellip.} Both have the same Vicon markers on the top face and Apriltag markers on the front face.}
  \label{fig:object_cropped_small}
\end{figure}

\myparagraph{Objects} 
We use two objects as shown in  \figref{fig:object_cropped_small}. The crucial part of each object is the bottom face, where the contact happens with the walls. Object \texttt{rect} has a rectangular face; \texttt{ellip} has an elliptical face, with these properties:
\begin{itemize}
\item \texttt{rect}: a 3D printed cuboid of $8\times5\times8$ cm, 110 g.
\item \texttt{ellip}: we modified the cuboid to have the bottom portion to be an elliptical cylinder. This is to maintain the same configuration on the top portion for sticking on the same pattern of Vicon markers and Apriltags as in \texttt{rect}. It weighs 100 g.

\end{itemize}

\myparagraph{Walls.}
Two walls have a top face of $4.5\times15.5$ cm. They are held rigidly to the environment.

\myparagraph{Robot.}
We use an ABB IRB 120 industrial robotic arm with 6 DOF to control precisely the position and velocity of its tool center point (TCP). The TCP moves at 5~mm/s in the experiments. The TCP pose is published at 250~Hz.

\myparagraph{Force sensing.}
We use an ATI Gamma F/T sensors, which connects the robot's TCP and the suction tube guide. It has high sensitivity with force resolution of 1/160 N, and torque resolution of 1/2000 N$\cdot$m. The data is published at 250~Hz.

\myparagraph{Vision system.} We use AprilTags to track the part at 30~Hz. We build a stereo AprilTag 3D pose tracking system to provide a more accurate and stable 3D visual tracking system than that with a single camera. We use two RGB cameras on two Intel Realsense SR300s to build a stereo system. 
The camera resolution used is $640\times480$ on both cameras.

\myparagraph{Suction system.} The suction cup is Piab BX25, which is made of a mix of rubber and polyurethane. The skirt of the diameter of the suction cup is 25mm. It is mounted on the suction tube guide. We use a vacuum generator that converts compressed air at 50 psi to vacuum.

\myparagraph{Data collection procedure.} We first command the robot to suction the object at the center, move the object above the hole, and align the object with the hole at the center. With the above as the nominal configuration, we add variations in translation in $x$ and rotation in yaw using the coordinate system in \figref{fig:setup}. We assume the error of the hole and object alignment is not large: within $15$mm in translation and {$15^\circ$} in rotation. 

We collect data for training and testing. The former is for learning the contact formation prediction and tuning the iSAM parameters.
For training the variations are {$\pm15$mm} and {$\pm 15^\circ$} with 25 grids evenly divided on both dimension; 625 trials in total, taking about 1 hour. For testing the variations are {$\pm14$mm} and {$\pm 14^\circ$} with 5 evenly divided grids in both dimension; i.e., 25 trials in total, taking about 2.5 minutes. The testing configurations are chosen so that they do not exactly match the same poses as in training in order to test generalization of the system. Each trial starts with a varied pose and then the robot moves straight down until the force sensor detects contact. After that, it moves straight up. 

The data is recorded using ROS Bag \cite{quigley2009ros} format so that we can test the same data with different system configurations.

\myparagraph{Groundtruth.}
We use a Vicon tracking system to accurately find the groundtruth pose of the object and the walls.
We also use the pose information to find the groundtruth of the CF labels by checking the geometric relationship between the bottom vertices of the object and the two line segments on the walls. This automatic labeling method allows us to label a large amount of data for training and testing.

\myparagraph{Computation.} All computation was done on a laptop machine with Intel Core i7-3920XM CPU and 16~GB RAM.

\begin{table}
  \caption{Contact formation prediction accuracy}
  \label{tab:svm_accuracy}
	\centering
	\begin{tabular}{|l|r|r|}
          \hline
          
          & \bf \texttt{rect} & \bf \texttt{ellip}     \\
          \hline
          {number of training data}    & 32,437    & 13,553  \\
           
          \hline 
          {cross-validation accuracy (\%)}  & 88.5      & 99.3  \\
          \hline
          
          {testing accuracy (\%)}  & 83.5      & 98.6 \\  
          
          \hline 
    \end{tabular}
\end{table}

\subsection{Accuracy of contact formation prediction}
We extract the portion of data where contact happens. \tabref{tab:svm_accuracy} reports the number of training samples and accuracy results. 

\begin{figure}
  \begin{center}
    \includegraphics[width=1\linewidth]{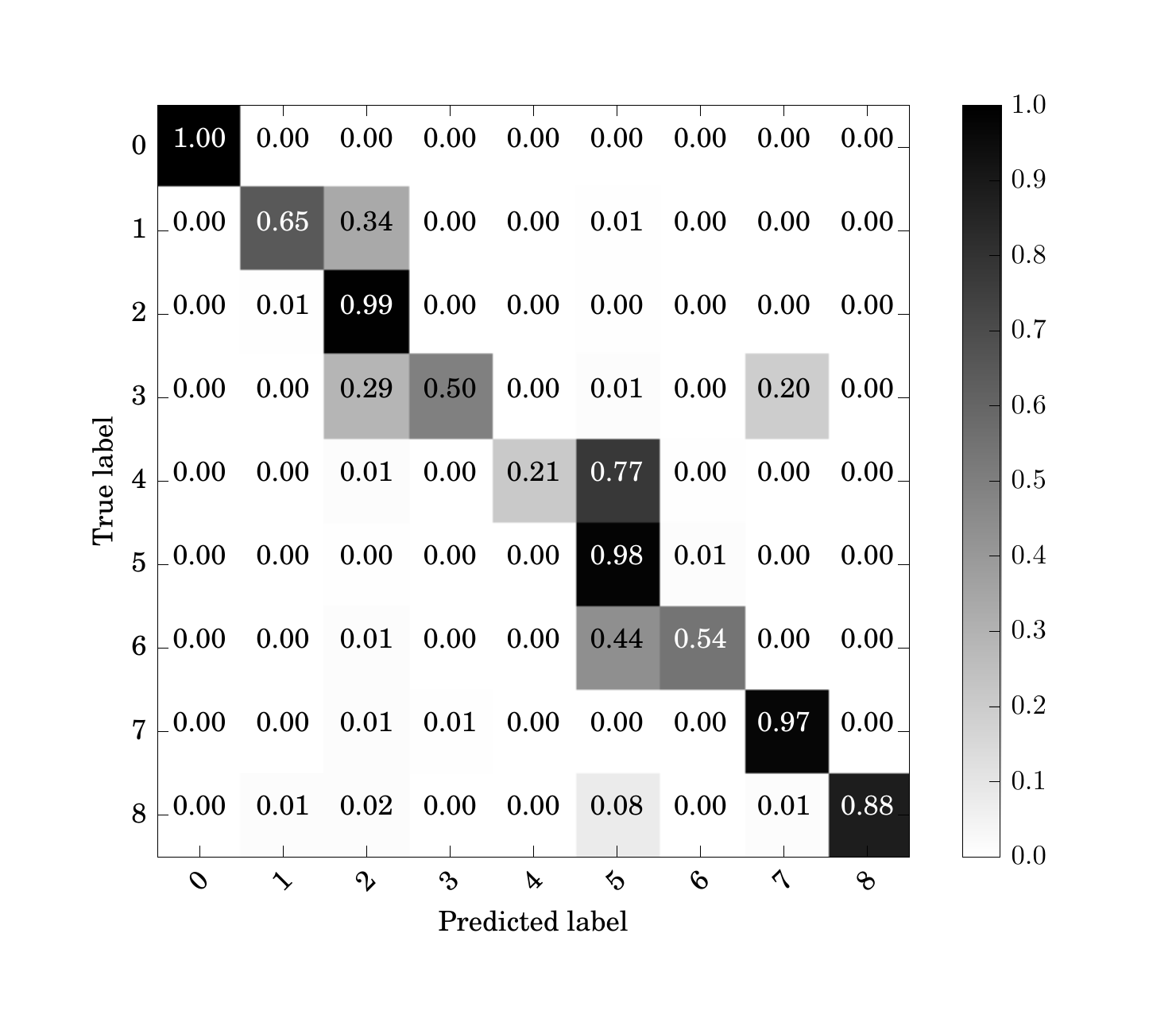}
    \includegraphics[width=0.6\linewidth]{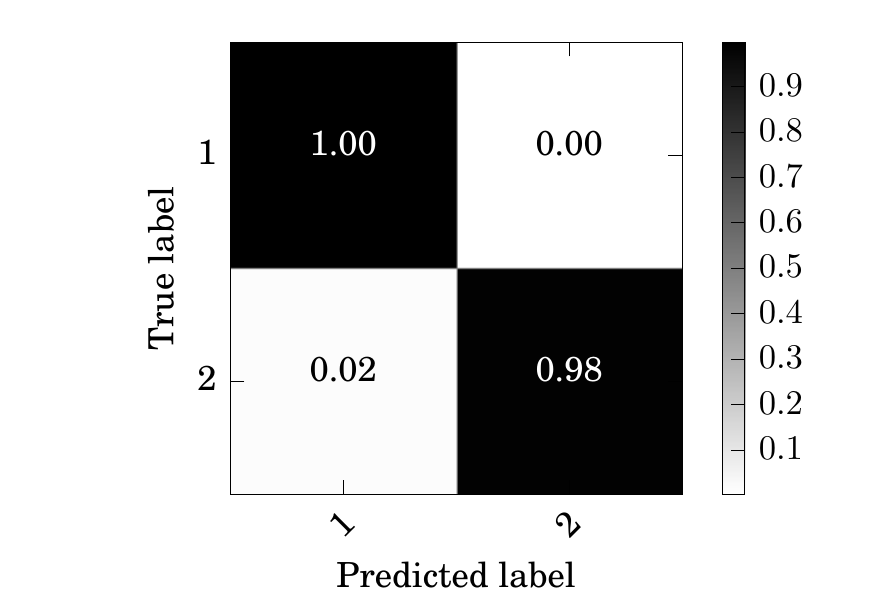}
  \end{center}
  \caption{Confusion matrix of SVM classification. (top) \texttt{rect}. (bottom) \texttt{ellip}}
  \label{fig:confusion_matrix}
\end{figure}

We examine the error types by checking the confusion matrices shown in \figref{fig:confusion_matrix}. For \texttt{rect}, notice the classification error is mostly among the contact formation from adjacent configurations, e.g., $\{1,2\}$, $\{2,3\}$, $\{4,5\},\{5,6\},\{3,7\}$ and $\{4,8\}$. Those kinds of error are usually not harmful to the state estimation, as the object pose is close for those CFs. We also note that there is more training data for the CF 2 and 5 case. Therefore, there is a bias toward those two classes. However, we do not intend to normalize the data because we assume the training distribution would be close to testing distribution. For \texttt{ellip}, we observe very little confusion between the two CFs. That suggests that \texttt{ellip} is much easier than \texttt{rect}.

\begin{figure}
  \begin{center}
    \includegraphics[width=0.9\linewidth]{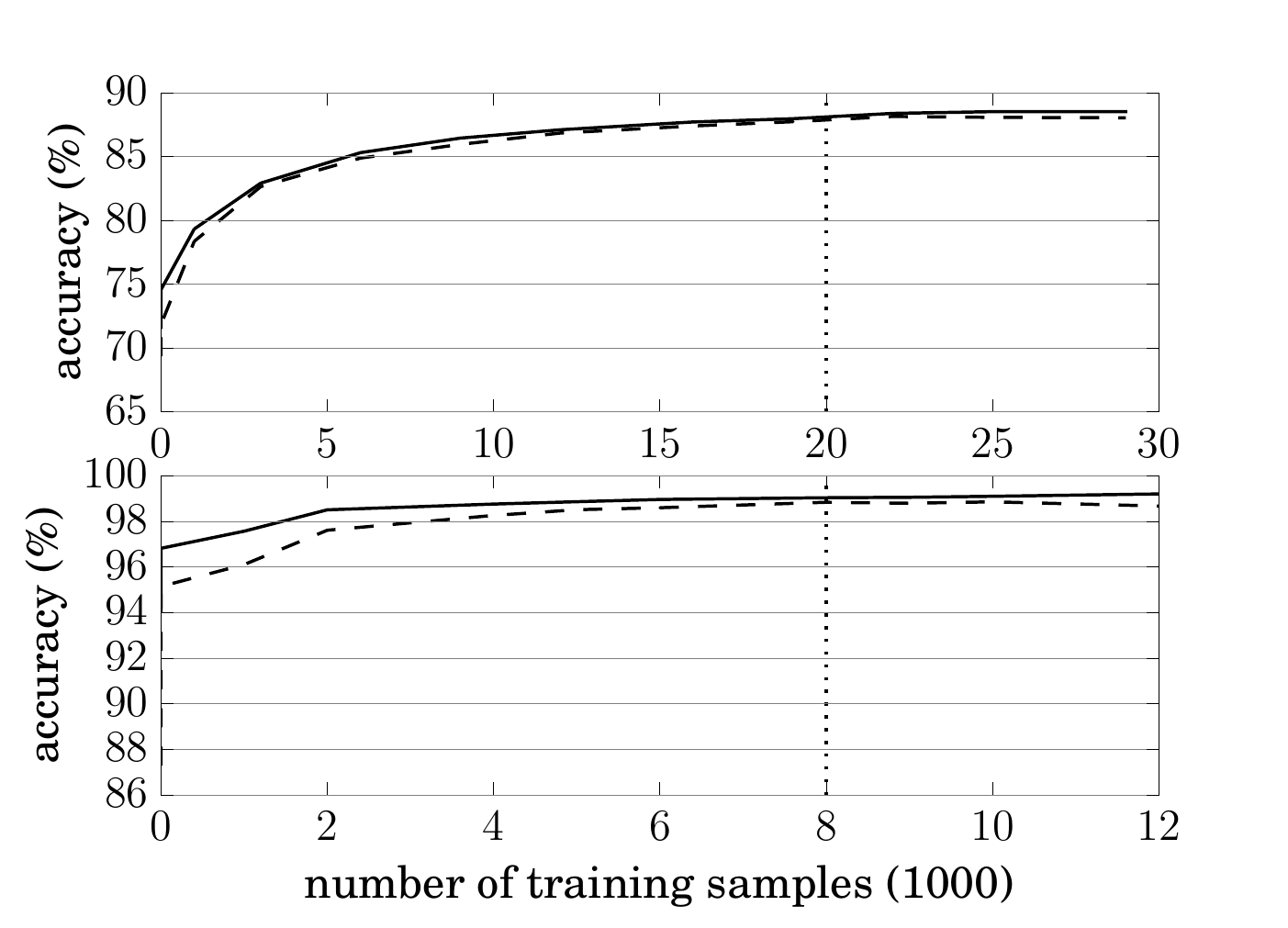}
  \end{center}
  \caption{Average (continuous) and worst (dashed) accuracy versus number of training samples. (top) \texttt{rect}. (bottom) \texttt{ellip}. The dotted lines represent saturation points.}
  \label{fig:accuracy_vs_numtrain}
\end{figure}

We also investigate how much training data is sufficient. \figref{fig:accuracy_vs_numtrain} shows the cross-validation accuracy versus the number of training samples. For \texttt{rect}, the accuracy increases with the amount of training samples until saturating at around 20,000; for \texttt{ellip}, it's around 8,000.

\subsection{Improvement of contact information}

We want to examine whether adding the proposed tactile costs -- contact constraint $C$ and robot motion $M$ -- would improve over the baseline, which uses only visual input. Here we use object \texttt{rect}.

\begin{table}
  \caption{RMSE with different combination of cost (object rect)}
  \label{tab:different_costs}
	\centering
	\begin{tabular}{|l|r|r|r|r|}
          \hline
          
          \bf Combination of costs & \bf Trans. & \bf Rot.  & \bf Trans(C). & \bf Rot(C). \\
          \bf       & \bf (mm)   & \bf (deg) & \bf (mm)      & \bf (deg)   \\ 
          \hline
           {vision (baseline)}          & 8.7$\pm$1.1 & 3.3$\pm$1.4 & 8.7$\pm$1.1 & 3.3$\pm$1.8  \\

          \hline
           {vision+robot} & 8.5$\pm$0.4 & 2.9$\pm$1.0 & 8.2$\pm$0.6 & 2.9$\pm$1.2 \\   
           
          \hline
           {vision+contact}  & 8.2$\pm$1.6 & 3.3$\pm$1.4 & 6.3$\pm$1.6 & 3.3$\pm$1.7 \\   
          
          \hline 
           {vision+contact+robot}  & 5.8$\pm$1.3 & 2.9$\pm$1.0 & 6.3$\pm$1.7 & 2.9$\pm$1.2 \\    
          \hline

    \end{tabular}
\end{table}

\begin{figure}
  \begin{center}
    \includegraphics[width=1\linewidth]{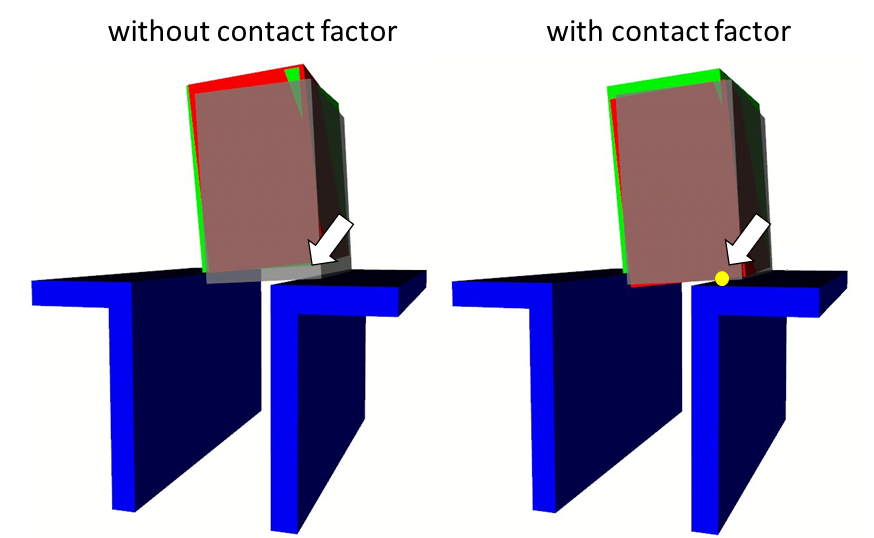}
    \includegraphics[width=1\linewidth]{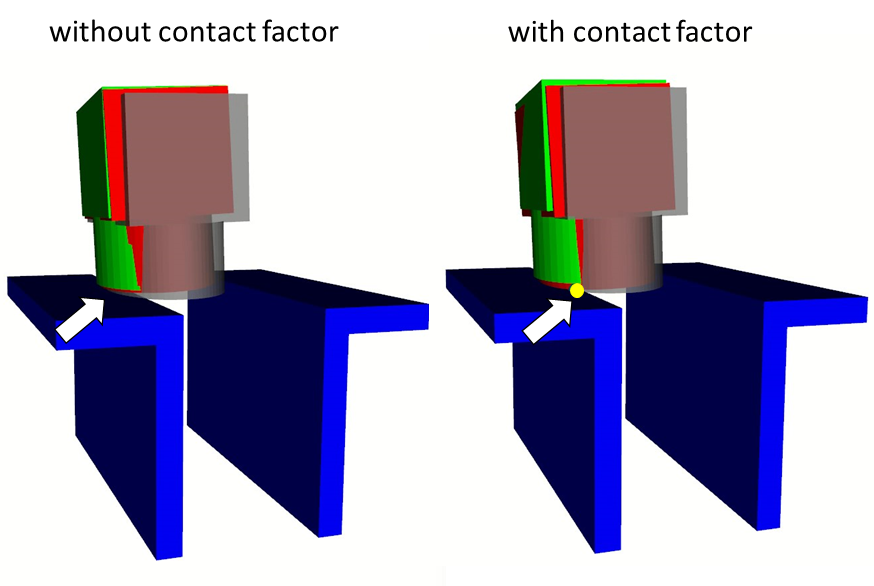}
  \end{center}
  \caption{Contact improvement example for \texttt{rect} (top) and \texttt{ellip} (bottom). Red: estimate. Gray: groundtruth. Green: vision input. Yellow dot: contact point. Notice where the contact happens, if there is no contact factor, there is a gap between the object and the wall but not in the configuration with contact factor. Best viewed in color.}
  \label{fig:contact_improve_example}
\end{figure}

We observe that both tactile-related constraints need to coexist in order to have a good improvement. \tabref{tab:different_costs} shows the root-mean-square errors in translation and rotation of the whole trajectory and errors only during contact, which are marked with (C).  For the first improvement, we add robot motion to the baseline. The result improves slightly: by 0.2~mm and 0.5~mm for full and contact portion, respectively. 
For the second improvement, we add contact information to the baseline. The result improves by 2.4~mm during the contact portions but not significant for the full trajectory. Combining both improvements, we see a larger 2.9~mm improvement on the whole trajectory. \figref{fig:contact_improve_example} shows on the top a time instance where the contact improves the accuracy.


\subsection{Evaluating contact point accuracy}

The groundtruth of detecting contact points is evaluated from Vicon's tracking result. We calculate the contact points by finding the closest point between the two line segments that are supposed to meet given the groundtruth CF. 

We see an improvement in contact point estimation as shown in \tabref{tab:contact_point}. We only report the portion where there is contact. We can see that adding robot motion information improves the accuracy by 5.1~mm, and adding contact information improves accuracy by 12.7~mm in error. With both additions, the accuracy improves by 13.6~mm and reaches an error of 5.4~mm. 

\begin{table} 
  \caption{RMSE of contact point estimation with different combination of cost for object rect }
  \label{tab:contact_point}
	\centering
	\begin{tabular}{|l|r|r|}
          \hline
          
          \bf Combination of costs & \bf Error (mm)    \\
           
          
          \hline
           {vision (baseline)} & 19.0$\pm$4.6      \\  
          \hline
          
           {vision+robot}    & 13.9$\pm$8.1    \\  
          \hline
           {vision+contact}      & 6.3$\pm$4.6  \\  
          \hline
           
           {vision+contact+robot} & 5.4$\pm$3.7      \\   
          \hline 
    \end{tabular}
\end{table}

\subsection{Different geometry of the object}
Here we show that our state estimation, using tactile sensing, can adapt to a different object shape. \figref{fig:contact_improve_example} bottom shows an instance of improvement using contact information. The contact constraint drags the estimation down to touch the wall when contact is detected. \tabref{tab:different_costs_ellipse} shows that adding the contact information improves the accuracy by 1.1~mm.

\begin{table}
  \caption{RMSE with object ellip }
  \label{tab:different_costs_ellipse}
	\centering
	\begin{tabular}{|l|r|r|r|r|}
          \hline
          
          \bf Combination of costs & \bf Trans. & \bf Rot.  & \bf Trans(C). & \bf Rot(C). \\
          \bf       & \bf (mm)   & \bf (deg) & \bf (mm)      & \bf (deg)   \\ 
          \hline
           {vision (baseline)}           & 8.0$\pm$1.0 & 2.9$\pm$1.2 & 8.0$\pm$1.0 & 2.9$\pm$1.1  \\

          \hline 
           {vision+robot+contact}  & 6.9$\pm$0.7 & 2.5$\pm$0.8 & 6.0$\pm$0.6 & 2.5$\pm$0.7 \\
  
          \hline

    \end{tabular}
\end{table}


\subsection{Run time}
The testing sequence is 154 sec long and the state estimator runs for 1,570 time steps, which results in 10~Hz or 96~ms per step in average. SVM prediction is fast: it takes 0.3~ms per step for \texttt{rect} and 0.06~ms per step for \texttt{ellip}, which indicates that the algorithm could run potentially faster.


\section{Conclusions}
In this work, we present a real-time state estimation system to recover the pose, the CF and contact points of an object interacting with its environment. We focus on the particular problem of inserting a suction-held object into a tight container. We use iSAM as the optimization framework to fuse tactile and visual sensing. At every timestep, we use an SVM classifier to decide the CF, and then add the corresponding contact constraints to iSAM. We show that using tactile sensing improves the accuracy of the estimated pose of the object with respect to visual sensing.

\myparagraph{Number of contact formations.} 
For the sake of computational complexity and to boost the performance of the SVM classifier, we use a reduced number of CFs. The CFs we chose are based on observation of real trials. There are two lessons: 1) If the CFs are too similar, merge them if possible; 2) If the CF happens rarely, regard it as no contact or the closest CF. 

\myparagraph{SVM accuracy.} 
We observe lower accuracy when we test the SVM on the test data. We believe this is in part due to the fact that the frequency of each formation is different during training and testing. During training, we want to preserve this prior knowledge of the frequency of different CFs so we did not normalize the data. Meanwhile, hardware degradation during the data collection process could also be a factor.

\myparagraph{Use force for locating contact points.} 
In this paper, we use a F/T sensor to find the discrete contact formation rather than finding the contact point location. Initially, we attempted to identify a dynamic model of the suction cup compliance to exploit this information jointly with the F/T readings to infer the contact location. In practice, the force reading is too noisy for this to yield a meaningful result.

\myparagraph{Unilateral Contact Constraints.}
While we can impose constraints of contact, we cannot guarantee that when there is no contact, objects will not penetrate. To prevent it, we would need unilateral constraints, which are not allowed by the optimization framework we propose.
%

\myparagraph{Future work.}
The CFs here are defined manually. One extension would be to investigate their automatic definition, as in \cite{tang2007representation} along with the contact constraints.
Based on the pose and contact state output, we would like to build a model-predictive controller for the insertion task by modifying the approach in \cite{hogan2016feedback} for pushing tasks.


\bibliographystyle{IEEEtranN} 
{\footnotesize \bibliography{state-est}} 

\end{document}